\documentclass{article}

\usepackage{times}
\usepackage{graphicx} % more modern
\usepackage{subfigure} 

\usepackage{natbib}

\usepackage{algorithm}
\usepackage{algorithmic}

\usepackage{hyperref}

\usepackage[accepted]{icml2017}

\icmltitlerunning{The Mythos of Model Interpretability}

\begin{document} 

\twocolumn[
\icmltitle{The Mythos of Model Interpretability}



% you can specify symbols, otherwise they are numbered in order
\icmlsetsymbol{equal}{*}

\begin{icmlauthorlist}
% \icmlauthor{Chris Donahue}{equal,to}
\icmlauthor{Zachary C. Lipton}{ucsd}
% \icmlauthor{Julian McAuley}{ucsd}
\end{icmlauthorlist}
\icmlaffiliation{ucsd}{University of California, San Diego}
\icmlcorrespondingauthor{Zachary C. Lipton}{zlipton@cs.ucsd.edu}
% \icmlcorrespondingauthor{Eee Pppp}{ep@eden.co.uk}

\icmlkeywords{Interpretability, Linear Models, Deep Learning}

\vskip 0.3in
]
\printAffiliationsAndNotice{} 
%\printAffiliationsAndNotice{\icmlEqualContribution} 

\begin{abstract} 
Supervised machine learning models 
boast remarkable predictive capabilities. 
But can you trust your model? 
Will it work in deployment? 
What else can it tell you about the world?
We want models to be not only good, but interpretable.
And yet the task of \emph{interpretation} appears underspecified.
Papers provide diverse and sometimes non-overlapping 
motivations for interpretability,
and offer myriad notions 
of what attributes render models interpretable.
Despite this ambiguity, 
many papers proclaim interpretability axiomatically, 
absent further explanation.
In this paper, we seek to refine the discourse on interpretability.
First, we examine the motivations
underlying interest in interpretability,
finding them to be diverse and occasionally discordant.
Then, we address model properties and techniques 
thought to confer interpretability,
identifying transparency to humans
and post-hoc explanations as competing notions.
Throughout, we discuss the feasibility 
and desirability of different notions, 
and question the oft-made assertions  
that linear models are interpretable
and that deep neural networks are not.
\end{abstract}

\section{Introduction}
\label{sec:introduction}
As machine learning models penetrate critical areas 
like medicine, the criminal justice system, and financial markets,
the inability of humans to understand these models seems problematic 
\cite{caruana2015intelligible, kim2015interactive}.
Some suggest \emph{model interpretability} as a remedy,
but few articulate precisely \emph{what} interpretability means 
or \emph{why} it is important.
Despite the absence of a definition,
papers frequently make claims about the interpretability of various models. 
From this, we might conclude that either: (i)
the definition of interpretability 
is universally agreed upon, but no one has managed to set it in writing,
or (ii) the term interpretability is ill-defined, 
and thus claims regarding interpretability of various models
may exhibit a quasi-scientific character.
Our investigation of the literature 
suggests the latter to be the case.
Both the motives for interpretability 
and the technical descriptions of interpretable models  
are diverse and occasionally discordant,
suggesting that interpretability 
refers to more than one concept. 
In this paper, we seek to clarify both,
suggesting that \emph{interpretability}
is not a monolithic concept, 
but in fact reflects several distinct ideas. 
We hope, through this critical analysis, 
to bring focus to the dialogue.

Here, we mainly consider supervised learning
and not other machine learning paradigms, 
such as reinforcement learning and interactive learning.
This scope derives from our original interest
in the oft-made claim that linear models are preferable 
to deep neural networks on account of their interpretability
\citep{lou2012intelligible}. 
To gain conceptual clarity,
we ask the refining questions: 
\emph{What is interpretability 
and why is it important?}
Broadening the scope of discussion seems counterproductive with respect to our aims.
For research investigating interpretability in the context of reinforcement learning, we point to \cite{dragan2013legibility} which studies the human interpretability of robot actions.
By the same reasoning, we do not delve as much as other papers might into Bayesian methods, however try to draw these connections where appropriate.

To ground any discussion of what might constitute interpretability, 
we first consider the various desiderata put forth in work addressing the topic (expanded in \S \ref{sec:desiderata}). 
Many papers propose interpretability
as a means to engender trust \citep{kim2015interactive, ridgeway1998interpretable}.
But what is trust? 
Does it refer to faith in a model's performance \cite{ribeiro2016should},
robustness, or to some other property of the decisions it makes?
Does interpretability simply mean a low-level mechanistic understanding of our models? 
If so does it apply to the features, parameters, models, or training algorithms?
Other papers suggest a connection between an interpretable model and one which uncovers causal structure in data \cite{athey2015machine}.
The legal notion of a \emph{right to explanation} offers yet another lens on interpretability. 

Often, our machine learning problem formulations are imperfect matches for the real-life tasks they are meant to solve.
This can happen when simplified optimization objectives 
fail to capture our more complex real-life goals.
Consider medical research with longitudinal data.
Our real goal may be to discover potentially causal associations, as with smoking and cancer \citep{wang1999smoking}. 
But the optimization objective for most supervised learning models is simply to minimize error, a feat that might be achieved in a purely correlative fashion.

Another such divergence of real-life and machine learning problem formulations emerges when the off-line training data for a supervised learner is not perfectly representative 
of the likely deployment environment.
For example, the environment is typically not stationary.
This is the case for product recommendation,
as new products are introduced 
and preferences for some items shift daily.
In more extreme cases, 
actions influenced by a model may alter the environment, invalidating future predictions.

Discussions of interpretability sometimes suggest 
that human decision-makers are themselves interpretable
because they can explain their actions \cite{ridgeway1998interpretable}.
But precisely what notion of interpretability 
do these explanations satisfy?
They seem unlikely to clarify the mechanisms 
or the precise algorithms by which brains work.
Nevertheless, the information conferred by an interpretation may be useful.
Thus, one purpose of interpretations 
may be to convey useful information of any kind.

After addressing the desiderata of interpretability, we consider what properties of models might 
render them interpretable (expanded in \S \ref{sec:properties}).
Some papers equate interpretability with \emph{understandability} 
or \emph{intelligibility} \citep{lou2013accurate},
i.e., that we can grasp \emph{how the models work}.
In these papers, understandable models 
are sometimes called \emph{transparent}, 
while incomprehensible models are called \emph{black boxes}. 
But what constitutes transparency?
We might look to the algorithm itself. 
Will it converge? Does it produce a unique solution?
Or we might look to its parameters: do we understand what each represents?
Alternatively, we could consider the model's complexity. 
Is it simple enough to be examined all at once by a human?

Other papers investigate so-called post-hoc interpretations.
These interpretations might \emph{explain} predictions without elucidating the mechanisms 
by which models work.
Examples of post-hoc interpretations
include the verbal explanations produced by people
or the saliency maps used to analyze deep neural networks.
Thus, humans decisions might admit post-hoc interpretability 
despite the \emph{black box} nature of human brains, 
revealing a contradiction between 
two popular notions of interpretability.

% In the following two sections, we expand ou
% The rest of this paper is organized as follows: 
% We examine motivations of interpretability 
% (\S \ref{sec:motivations}). 
% We explore model properties considered to confer interpretability (\S \ref{sec:properties}). 
% We discuss the feasibility and desirability 
% of various notions of interpretability (\S \ref{sec:discussion}). 

\section{Desiderata of Interpretability Research}
\label{sec:desiderata}
At present, \emph{interpretability} has no formal technical meaning.
One aim of this paper is to propose more specific definitions.
Before we can determine which meanings might be appropriate,
we must ask what the real-world objectives of interpretability research are.
In this section we spell out the various  desiderata of interpretability research 
through the lens of the literature. 

While these desiderata are diverse,
it might be instructive to first consider a common thread that persists throughout the literature:
The demand for interpretability arises when there is a mismatch between the formal objectives of supervised learning (test set predictive performance) and the real world costs in a deployment setting. 

\begin{figure}[ht!]
	\centering
	\includegraphics[scale=.35]{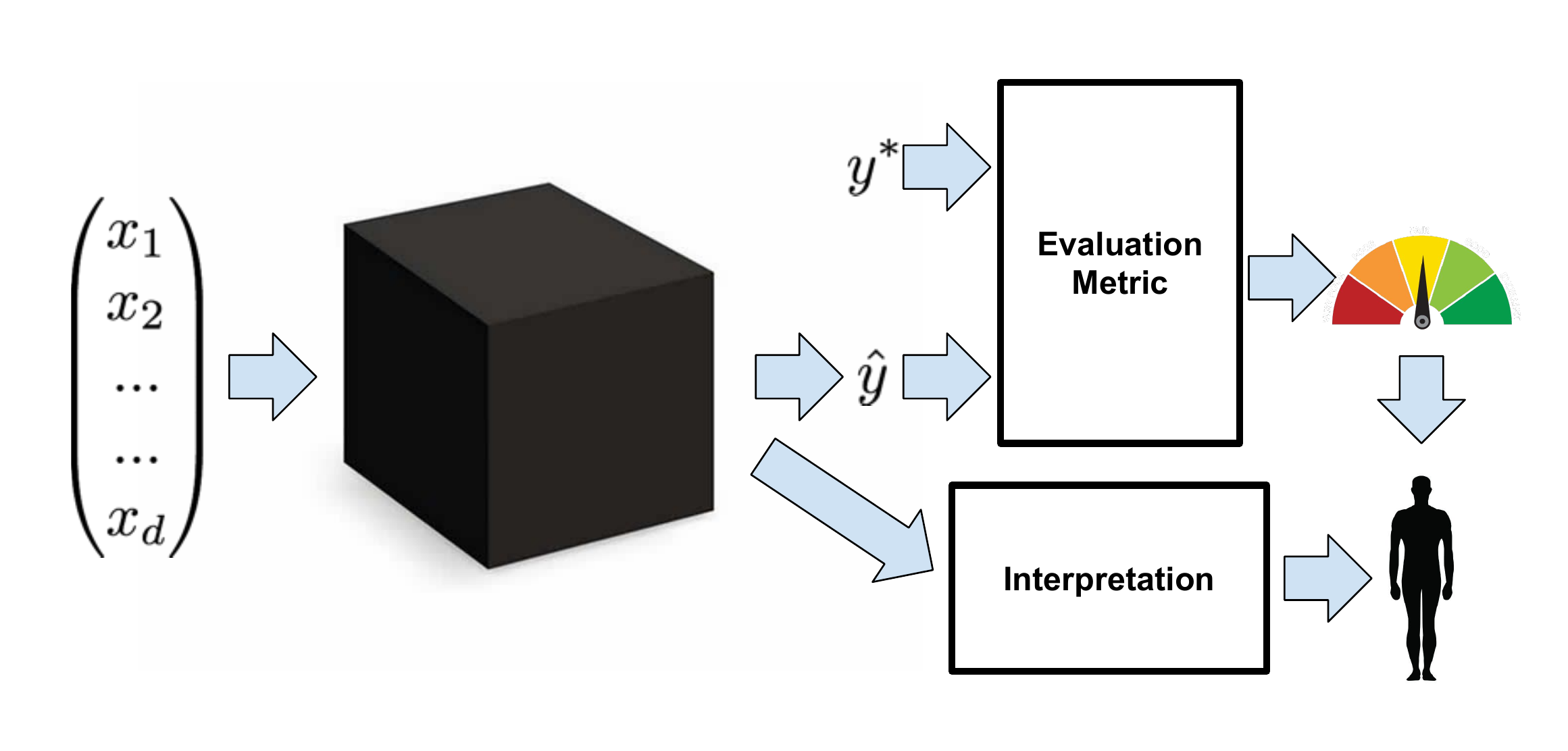}
    \caption{Typically, evaluation metrics require only predictions and \emph{ground truth} labels. When stakeholders additionally demand \emph{interpretability}, we might infer 
the existence of desiderata that cannot be captured in this fashion.}
\label{fig:interpretability-flow}
\end{figure}

Consider that most common evaluation metrics for supervised learning 
require only predictions, together with ground truth, to produce a score.
These metrics can be be assessed for every supervised learning model.
So the very desire for an \emph{interpretation}
suggests that in some scenarios, predictions alone 
and metrics calculated on these predictions do not suffice to characterize the model (Figure \ref{fig:interpretability-flow}).
% This implies a discrepancy between our real-world objectives and the simple objectives 
% optimized by machine learning models during training.
We should then ask, what are these other desiderata and under what circumstances are they sought? 
% Under what circumstances do metrics like accuracy, area under the receiver operating characteristic, and cross entropy become weak surrogates for the real-world goals of machine learning practitioners?

However inconveniently, 
it turns out that many situations arise 
when our real world objectives 
are difficult to encode as simple real-valued functions.
For example, an algorithm for making hiring decisions 
should simultaneously optimize productivity, ethics, and legality.
But typically, ethics and legality cannot be directly optimized.
The problem can also arise when the dynamics of the deployment environment differ from the training environment.
In all cases, \emph{interpretations} serve those objectives that we deem important but struggle to model formally.

\subsection{Trust}
Some papers motivate interpretability 
by suggesting it to be prerequisite for \emph{trust} \citep{kim2015interactive,ribeiro2016should}.
But what is trust?
Is it simply confidence 
that a model will perform well?
If so, a sufficiently accurate model
should be demonstrably trustworthy
and interpretability would serve no purpose.
Trust might also be defined subjectively.
For example, a person might feel more at ease 
with a well-understood model, 
even if this understanding served no obvious purpose. 
Alternatively, when the training and deployment objectives diverge, 
trust might denote confidence 
that the model will perform well 
with respect to the real objectives and scenarios.
% For example, in the case of self-driving cars, 
% we might not trust them to respond appropriately 
% in unforeseen situations.

For example, 
consider the growing use of machine learning models to forecast crime rates for purposes of allocating police officers.
We may trust the model to make accurate predictions
but not to account for racial biases in the training data for the model's own effect in perpetuating a cycle of incarceration by over-policing some neighborhoods.
Another sense in which we might trust a machine learning model might be 
that we feel comfortable relinquishing control to it.
In this sense, we might care not only about \emph{how often a model is right} 
but also \emph{for which examples it is right}. 
If the model tends to make mistakes in regions of input space 
where humans also make mistakes, 
and is typically accurate when humans are accurate, 
then it may be considered trustworthy 
in the sense that there is no expected cost 
of relinquishing  control. 
But if a model tends to make mistakes for inputs that humans classify accurately, then there may always be an advantage to maintaining human supervision of the algorithms. 

\subsection{Causality}
Although supervised learning models
are only optimized directly to make associations,
researchers often use them
in the hope of inferring properties 
or generating hypotheses about the natural world.
For example, a simple regression model 
might reveal a strong association 
between thalidomide use and birth defects or smoking and lung cancer \cite{wang1999smoking}.

The associations learned by supervised learning algorithms are not guaranteed to reflect causal relationships. 
There could always exist unobserved causes 
responsible for both associated variables.
One might hope, however, 
that by interpreting supervised learning models,
we could generate hypotheses that scientists could then test experimentally. 
\citet{liu2005empirical}, for example, emphasizes regression trees and Bayesian neural networks, suggesting that models are interpretable 
and thus better able to provide clues 
about the causal relationships 
between physiologic signals and affective states. 
The task of inferring causal relationships from observational data has been extensively studied \citep{pearl2009causality}. 
But these methods tend to rely on strong assumptions of prior knowledge.
% Among ad-hoc approaches to data-mining, there is no consensus
% However, on real-world data, 
% among purely observational and correlative machine learning methods, which ones tend to yield the most promising hypotheses of causal relationships remains an open question.

\subsection{Transferability} 
Typically we choose training and test data 
by randomly partitioning examples from the same distribution.
We then judge a model's generalization error 
by the gap between its performance 
on training and test data.
However, humans exhibit a far richer capacity to generalize, 
transferring learned skills to unfamiliar situations.
We already use machine learning algorithms in situations where such abilities are required,
such as when the environment is non-stationary.
We also deploy models in settings 
where their use might alter the environment,
invalidating their future predictions. 
Along these lines, \citet{caruana2015intelligible}
describe a model trained 
to predict probability of death from pneumonia
that assigned less risk to patients 
if they also had asthma.
In fact, asthma was predictive of lower risk of death.
This owed to the more aggressive treatment 
these patients received.
But if the model were deployed to aid in triage, 
these patients would then receive less aggressive treatment, invalidating the model.

Even worse, we could imagine situations, like machine learning for security, where the environment might be actively adversarial.
Consider the recently discovered susceptibility of convolutional neural networks (CNNs) to adversarial examples.
The CNNs were made to misclassify images 
that were imperceptibly (to a human) perturbed \citep{szegedy2013intriguing}.
Of course, this isn't overfitting 
in the classical sense. 
The results achieved on training data 
generalize well to i.i.d.~test data.
But these are mistakes a human wouldn't make 
and we would prefer models not to make these mistakes either.

%
%	Real-world adversarial situation
%
Already, supervised learning models are regularly subject to such adversarial manipulation.
Consider the models used to generate credit ratings,
scores that when higher should signify a higher probability 
that an individual repays a loan. 
According to their own technical report, 
FICO trains credit models using logistic regression \citep{fico2011},
specifically citing interpretability 
as a motivation for the choice of model.
Features include dummy variables representing binned values for average age of accounts, debt ratio, and the number of late payments,
and the number of accounts in good standing.

Several of these factors can be manipulated at will by credit-seekers.
For example, one's debt ratio 
can be improved simply by requesting periodic increases to credit lines 
while keeping spending patterns constant.
Similarly, the total number of accounts 
can be increased by simply applying for new accounts, 
when the probability of acceptance is reasonably high. 
Indeed, FICO and Experian both acknowledge that credit ratings 
can be manipulated, 
even suggesting guides for improving one's credit rating. 
These rating improvement strategies 
do not fundamentally change one's underlying ability to pay a debt.
The fact that individuals actively and successfully game the rating system may invalidate its predictive power.

\subsection{Informativeness} 
Sometimes we apply decision theory to the outputs of supervised models to take actions in the real world.
However, in another common use paradigm,
the supervised model is used instead to provide
information to human decision makers,
a setting considered by \citet{kim2015ibcm, huysmans2011empirical}.
While the machine learning objective might be to reduce error,
the real-world purpose is to provide useful information. 
The most obvious way that a model conveys information is via its outputs. 
However, it may be possible via some procedure to convey additional information to the human decision-maker.

By analogy, we might consider a PhD student seeking advice from her advisor. 
Suppose the student asks what venue would best suit a paper. 
The advisor could simply name one conference,
but this may not be especially useful.
Even if the advisor is reasonably intelligent, 
the terse reply doesn't enable the student to meaningfully combine the advisor's knowledge with her own.

An interpretation may prove informative 
even without shedding light on a model's inner workings.
For example, a diagnosis model  
might provide intuition to a human decision-maker
by pointing to similar cases 
in support of a diagnostic decision.
In some cases, we train a supervised learning model, 
but our real task more closely resembles unsupervised learning.
Here, our real goal is to explore the data
and the objective serves only as \emph{weak supervision}.

\subsection{Fair and Ethical Decision-Making}
At present, politicians, journalists and researchers have expressed concern that we must produce \emph{interpretations} 
for the purpose of assessing whether decisions produced automatically by algorithms conform to ethical standards \citep{goodman2016european}. 

The concerns are timely: Algorithmic decision-making mediates more and more of our interactions, influencing our social experiences, the news we see, our finances, and our career opportunities. We task computer programs with approving lines of credit, curating news, and filtering job applicants. Courts even deploy computerized algorithms to predict “risk of recidivism”, the probability that an individual relapses into criminal behavior \citep{chouldechova2016fair}. It seems likely that this trend will only accelerate as breakthroughs in artificial intelligence rapidly broaden the capabilities of software.

Recidivism predictions are already used to determine who to release 
and who to detain, raising ethical concerns.~\footnote{
It seems reasonable to argue that under most circumstances,
risk-based punishment is fundamentally unethical, 
but this discussion requires exceeds the present scope.
}
How can we be sure that predictions do not discriminate 
on the basis of race?
Conventional evaluation metrics such as accuracy or AUC offer little assurance that a model and via decision theory, its actions, behave acceptably.
Thus demands for fairness often lead to demands for \emph{interpretable} models.

New regulations in the European Union propose that individuals affected by algorithmic decisions have a \emph{right to explanation} \citep{goodman2016european}.
Precisely what form such an explanation might take or how such an explanation could be proven correct and not merely appeasing remain open questions.
Moreover the same regulations suggest that  algorithmic decisions should be \emph{contestable}. So in order for such explanations to be useful it seems they must (i) present clear reasoning based on falsifiable propositions and (ii) offer some natural way of contesting these propositions and modifying the decisions appropriately if they are falsified.

%
% 	Section: Properties of Interpretable Models 
%
\section{Properties of Interpretable Models}
\label{sec:properties}
We turn now to consider the techniques and model properties
that are proposed either to enable or to comprise \emph{interpretations}.
These broadly fall into two categories.
The first relates to \emph{transparency}, i.e., \emph{how does the model work?}
The second consists of \emph{post-hoc explanations}, i.e., \emph{what else can the model tell me?} 
This division is a useful organizationally, 
but we note that it is not absolute. 
For example post-hoc analysis techniques attempt to uncover the significance of various of parameters, 
an aim we group under the heading of transparency.

\subsection{Transparency}
Informally, \emph{transparency} is 
the opposite of \emph{opacity} or \emph{blackbox-ness}. 
It connotes some sense of understanding the mechanism by which the model works.
We consider transparency 
at the level of the entire model (\emph{simulatability}),
at the level of individual components (e.g. parameters) (\emph{decomposability}),
and at the level of the training algorithm (\emph{algorithmic transparency}).

\subsubsection{Simulatability}
In the strictest sense, we might call a model transparent 
if a person can contemplate the entire model at once. 
This definition suggests that an interpretable model is a simple model.
We might think, for example that 
for a model to be fully understood, 
a human should be able to take the input data 
together with the parameters of the model 
and in \emph{reasonable} time step through every calculation 
required to produce a prediction. 
This accords with the common claim that sparse linear models, 
as produced by lasso regression \citep{tibshirani1996regression},
are more interpretable than dense linear models learned on the same inputs.
\citet{ribeiro2016should} also adopt this notion of interpretability,
suggesting that an interpretable model is one that ``can be readily presented to the user with visual or textual artifacts.''

For some models, such as decision trees, 
the size of the model (total number of nodes) 
may grow much faster than the time to perform inference (length of pass from root to leaf).
This suggests that simulatability may admit two subtypes, 
one based on the total size of the model 
and another based on the computation required to perform inference.

Fixing a notion of simulatability, 
the quantity denoted by \emph{reasonable} is subjective.
But clearly, given the limited capacity of human cognition, this ambiguity might only span several orders of magnitude. 
% Few would suggest that any normal human 
% could contemplate a thousand parameters at once, 
% but most could step through, or even memorize a linear model with 10 parameters. 
In this light, we suggest that neither linear models, rule-based systems, 
nor decision trees are intrinsically interpretable.
Sufficiently high-dimensional models, unwieldy rule lists, and deep decision trees could all be considered less transparent 
than comparatively compact neural networks.

\subsubsection{Decomposability} 
A second notion of transparency
might be that each part of the model -
each input, parameter, and calculation - 
admits an intuitive explanation.
This accords with the property of \emph{intelligibility}
as described by \cite{lou2012intelligible}.
For example, each node in a decision tree might correspond to a plain text description (e.g.~\emph{all patients with diastolic blood pressure over 150}).
Similarly, the parameters of a linear model could be described as representing strengths of association 
between each feature and the label.

Note that this notion of interpretability 
requires that inputs themselves be individually interpretable,
disqualifying some models 
with highly engineered or anonymous features.
While this notion is popular, 
we shouldn't accept it blindly. 
The weights of a linear model might seem intuitive, 
but they can be fragile with respect to feature selection and pre-processing. 
For example, associations between flu risk and vaccination 
might be positive or negative depending on whether the feature set 
includes indicators of old age, infancy, or immunodeficiency.

% Some Bayesian methods also exhibit transparency in the sense of decomposability.
% For example, probabilistic graphical models (PGMs) 
% express distributions with graphs that encode probabilistic relationships between events.
% The conditional probabilities relating events
% in a PGM admit clear intuitive explanations.
% For this reason a growing body of work addressing interpretability favors these methods \cite{kim2014bayesian, doshi2014graph, kim2015mind}.
% However, the decomposability of PGMs 
% can be compromised for sufficiently flexible 
% latent variable models \cite{chang2009reading}.

\subsubsection{Algorithmic Transparency}
A final notion of transparency
might apply at the level of the learning algorithm itself.
% Even absent the ability to mentally simulate a model
% or to intuit the meaning of its components.
For example, in the case of linear models, 
we understand the shape of the error surface.
We can prove that training will converge to a unique solution,
even for previously unseen datasets. 
This may give some confidence that the model might behave in an online setting requiring programmatic retraining on previously unseen data.
On the other hand, modern deep learning methods 
lack this sort of algorithmic transparency.
While the heuristic optimization procedures for neural networks are demonstrably powerful, we don't understand how they work, and at present cannot guarantee a priori that they will work on new problems.
Note, however, that humans exhibit none of these forms of transparency.

\subsection{Post-hoc Interpretability}
Post-hoc interpretability presents a distinct approach to extracting information from learned models.
% A distinct model-based notion of \emph{interpretability}, 
While post-hoc interpretations often 
do not elucidate precisely how a model works,
they may nonetheless confer useful information for practitioners and end users of machine learning. 
Some common approaches to post-hoc interpretations include natural language explanations, 
visualizations of learned representations or models, 
and explanations by example 
(e.g. \emph{this tumor is classified as malignant 
because to the model it looks a lot like these other tumors}).

To the extent that we might consider humans to be interpretable,
it is this sort of interpretability that applies.
For all we know, the processes by which we humans make decisions
and those by which we explain them may be distinct. 
One advantage of this concept of interpretability 
is that we can interpret opaque models after-the-fact,
without sacrificing predictive performance.

\subsubsection{Text Explanations}
% An often cited motivation for interpretability is that medical patients won't trust the decisions of a machine (or doctor) absent explanations. Some believe this trust to be essential for patient compliance \cite{carenini1994generating}.
% Accordingly, 
Humans often justify decisions verbally.
Similarly, we might train one model to generate predictions and a separate model, 
such as a recurrent neural network language model, 
to generate an explanation. 
Such an approach is taken in a line of work by \citet{krening2016learning}.
They propose a system in which one model (a reinforcement learner)
chooses actions to optimize cumulative discounted return.
They train another model to map a model's state representation onto 
verbal explanations of strategy.
These explanations are trained to maximize the likelihood of previously observed ground truth explanations from human players, 
and may not faithfully describe the agent's decisions, 
however plausible they appear.
We note a connection between this approach and recent work on neural image captioning in which the representations learned by a discriminative convolutional neural network (trained for image classification) 
are co-opted by a second model to generate captions.
These captions might be regarded as interpretations 
that accompany classifications.

In work on recommender systems, 
\citet{mcauley2013hidden} use text 
to explain the decisions of a latent factor model.
Their method consists of simultaneously training a latent factor model for rating prediction and a topic model for product reviews.
During training they alternate between decreasing the squared error on rating prediction and increasing the likelihood of review text.
The models are connected because they use normalized latent factors as topic distributions. 
In other words, latent factors are regularized such that they are also good at explaining the topic distributions in review text.
The authors then explain user-item compatibility by 
examining the top words in the topics corresponding to matching components of their latent factors.
% Note that these interpretation is not fully post-hoc 
% because the topic model is incorporated 
% into the training procedure.
Note that the practice of interpreting topic models by presenting the top words is itself a post-hoc interpretation technique that has invited scrutiny \citep{chang2009reading}.

\subsubsection{Visualization}
Another common approach to generating post-hoc interpretations is to render visualizations
in the hope of determining qualitatively 
what a model has learned. 
One popular approach is to visualize 
high-dimensional distributed representations 
with t-SNE \citep{van2008visualizing}, a technique that  renders 2D visualizations in which nearby data points are likely to appear close together.

\citet{mordvintsev2015inceptionism} attempt to explain 
what an image classification network has learned 
by altering the input through gradient descent to enhance the activations of certain nodes selected from the hidden layers.
An inspection of the perturbed inputs can give clues 
to what the model has learned.
Likely because the model was trained on a large corpus of animal images, they observed that enhancing some nodes caused the dog faces to appear throughout the input image.

In the computer vision community, similar approaches have been explored to investigate what information is retained at various layers of a neural network.
\citet{mahendran2015understanding} pass an image through a discriminative convolutional neural network to generate a representation.
They then demonstrate that the original image can be recovered with high fidelity even from reasonably high-level representations (level 6 of an AlexNet) 
by performing gradient descent 
on randomly initialized pixels.

\subsubsection{Local Explanations}
\begin{figure}[ht]
	\centering
	\includegraphics[width=\linewidth]{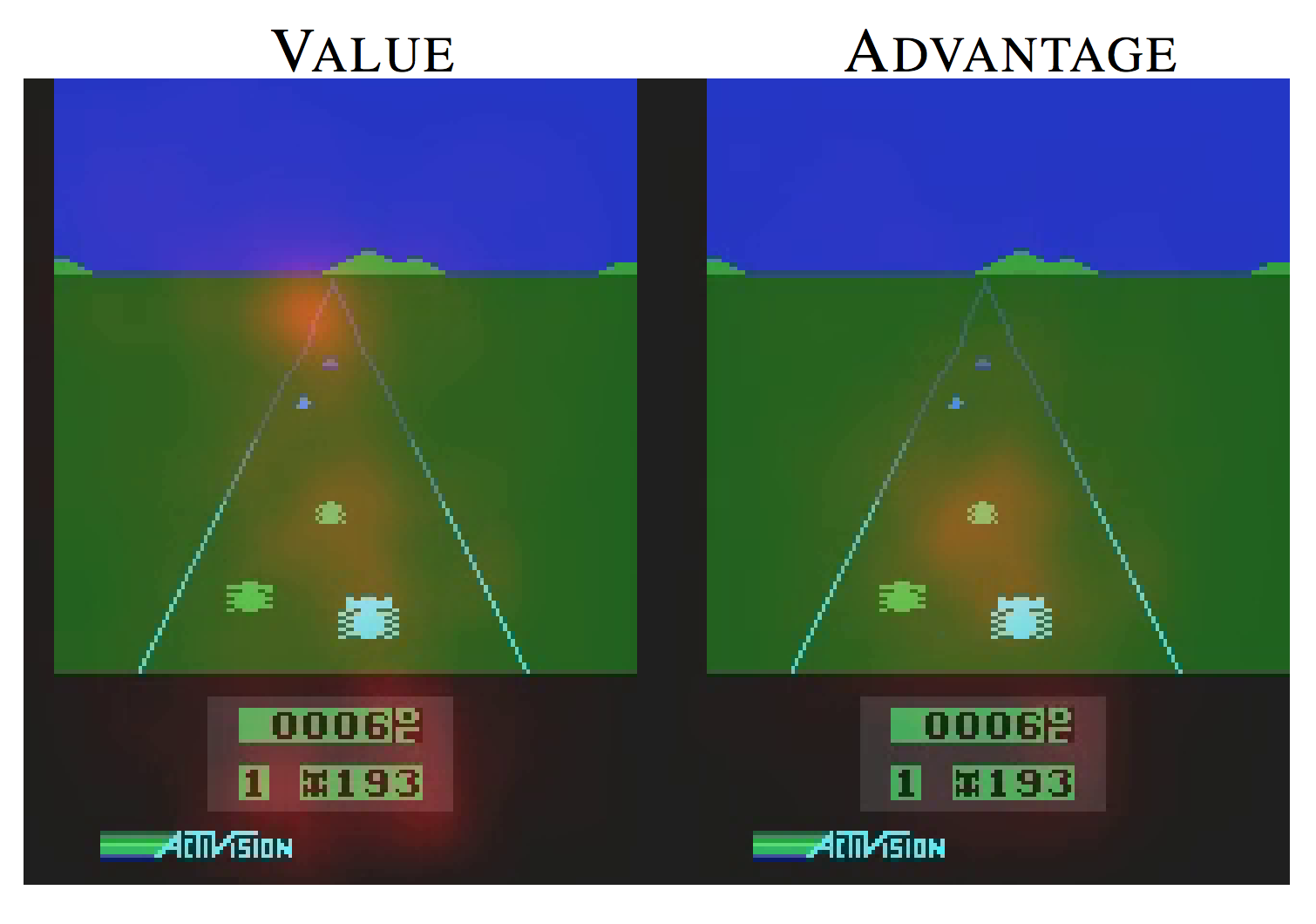}
    \caption{Saliency map by \citet{wang2015dueling} to convey intuition over what the value function and advantage function portions of their deep Q-network are \emph{focusing} on.}
\label{fig:saliency}
\end{figure}

While it may be difficult to succinctly describe the full mapping learned by a neural network, 
some papers focus instead on explaining 
what a neural network depends on locally.
One popular approach for deep neural nets
is to compute a saliency map.
Typically, they take the gradient of 
the output corresponding to the correct class 
with respect to a given input vector. 
For images, this gradient can be applied as a mask (Figure \ref{fig:saliency}), 
highlighting regions of the input that, if changed, would most influence the output \citep{simonyan2013deep,wang2015dueling}.

Note that these explanations 
of what a model is \emph{focusing on} may be misleading.
The saliency map is a local explanation only. 
Once you move a single pixel, 
you may get a very different saliency map.
This contrasts with linear models, 
which model global relationships between inputs and outputs.

Another attempt at local explanations is made by \citet{ribeiro2016should}.
In this work, the authors explain the decisions of any model in a local region near a particular point, by learning a separate sparse linear model to explain the decisions of the first.

\subsubsection{Explanation by Example} 
One post-hoc mechanism for explaining the decisions of a model 
might be to report (in addition to predictions)
which other examples the model considers to be most similar,
a method suggested by \citet{caruana1999case}.
After training a deep neural network or latent variable model for a discriminative task,
we then have access not only to predictions but also to the learned representations.
Then, for any example, in addition to generating a prediction, 
we can use the activations of the hidden layers 
to identify the $k$-nearest neighbors 
based on the proximity in the space learned by the model. 
This sort of explanation by example has precedent in how humans sometimes justify actions by analogy. % citations
For example, doctors often refer to case studies 
to support a planned treatment protocol.

In the neural network literature,  \citet{mikolov2013distributed} use such an approach to examine the learned representations of words after word2vec training.  While their model is trained for discriminative skip-gram prediction, 
to examine what relationships the model has learned, they enumerate nearest neighbors of words based on distances calculated in the latent space.
We also point to related work in Bayesian methods: \citet{kim2014bayesian} and \citet{doshi2014graph} investigate cased-base reasoning approaches for interpreting generative models.

% \section{Comparison of Deep Nets \& Linear Models}
% \input{linear_vs_deep.tex}

\section{Discussion}
\label{sec:discussion}
The concept of interpretability appears
simultaneously important and slippery.
Earlier, we analyzed both the motivations for interpretability and some attempts by the research community to confer it.
In this discussion, we consider the implications of our analysis and offer several takeaways to the reader.
 
\subsection{Linear models are not strictly more interpretable than deep neural networks}
Despite this claim's enduring popularity, its truth content
varies depending on what notion of interpretability we employ.
With respect to \emph{algorithmic transparency}, 
this claim seems uncontroversial, 
but given high dimensional or heavily engineered features, 
linear models lose \emph{simulatability} 
or \emph{decomposability}, respectively. 

When choosing between linear and deep models, 
we must often make a trade-off 
between \emph{algorithic transparency} and \emph{decomposability}.
This is because deep neural networks tend to operate on raw or lightly processed features. So if nothing else, the features are intuitively meaningful, and post-hoc reasoning is sensible.
However, in order to get comparable performance, 
linear models often must operate on heavily hand-engineered features.
\citet{lipton2016modeling} demonstrates such a case where linear models can only approach the performance of RNNs at the cost of decomposability. 

For some kinds of post-hoc interpretation, 
deep neural networks exhibit a clear advantage.
They learn rich representations 
that can be visualized, verbalized, or used for clustering. 
Considering the desiderata for interpretability,
linear models appear to have a better track record 
for studying the natural world 
% and for identifying weaknesses in training data,
but we do not know of a theoretical reason 
why this must be so.
Conceivably, post-hoc interpretations 
could prove useful in similar scenarios.

\subsection{Claims about interpretability must be qualified}
As demonstrated in this paper, 
the term does not reference a monolithic concept.
To be meaningful, any assertion regarding interpretability
should fix a specific definition. 
If the model satisfies a form of transparency, 
this can be shown directly.
For post-hoc interpretability, 
papers ought to fix a clear objective and demonstrate evidence 
that the offered form of interpretation achieves it.

\subsection{In some cases, transparency may be at odds with the broader objectives of AI}
Some arguments against \emph{black-box} algorithms
appear to preclude any model 
that could match or surpass our abilities on complex tasks. 
As a concrete example, 
the short-term goal of building trust with doctors 
by developing transparent models  
might clash with the longer-term goal 
of improving health care.
We should be careful when giving up predictive power, 
that the desire for transparency is justified 
and isn't simply a concession 
to institutional biases against new methods.

\subsection{Post-hoc interpretations can potentially mislead}
We caution against blindly embracing post-hoc notions of interpretability,
especially when optimized to placate subjective demands. 
In such cases, one might - deliberately or not - 
optimize an algorithm 
to present misleading but plausible explanations. 
As humans, we are known to engage in this behavior,
as evidenced in hiring practices and college admissions. 
Several journalists and social scientists 
have demonstrated that acceptance decisions attributed 
to virtues like \emph{leadership} or \emph{originality} 
often disguise racial or gender discrimination \citep{admissions}.
In the rush to gain acceptance for machine learning
and to emulate human intelligence, 
we should be careful 
not to reproduce pathological behavior at scale.

\subsection{Future Work}
We see several promising directions for future work.
First, for some problems, the discrepancy between real-life and machine learning objectives could be mitigated 
by developing richer loss functions and performance metrics. 
Exemplars of this direction include research on sparsity-inducing regularizers and cost-sensitive learning.
Second, we can expand this analysis 
to other ML paradigms such as reinforcement learning.
Reinforcement learners can address some (but not all)
of the objectives of interpretability research 
by directly modeling interaction between models and environments.
However, this capability may come at the cost of allowing models to experiment in the world, incurring real consequences.
Notably, reinforcement learners are able to learn causal relationships between their actions and real world impacts. 
However, like supervised learning, reinforcement learning relies on  a well-defined scalar objective.
For problems like fairness, where we struggle to verbalize precise definitions of success, a shift of ML paradigm is unlikely to eliminate the problems we face.

\section{Contributions}
\label{sec:contributions}
This paper identifies an important but under-recognized problem: 
the term \emph{interpretability} holds no agreed upon meaning, 
and yet machine learning conferences frequently publish papers which wield the term in a quasi-mathematical way. 
For these papers to be meaningful and for this field to progress, 
we must critically engage the issue of problem formulation.
Moreover, we identify the incompatibility of most presently investigated interpretability techniques with pressing problems facing machine learning in the wild. 
For example, little in the published work on model intepretability 
addresses the idea of contestability.
% And while the ability to fool a humans with plausible but untrue text explanations is academically interesting, 
% it does little to address
% And while a growing body of work addresses theoretical definitions fairness,
% currently available interpretations yielded by transparent models  post-hoc techniques do little to address the problem

This paper makes a first step towards providing a comprehensive taxonomy of both the desiderata and methods in interpretability research.
% We hope that the draft is evaluated on the strength and impact of its arguments and not by its conformity to an expected aesthetic. 
% We note that this paper does not introduce a new algorithm, 
% new mathematics, or experimental findings. 
% And we recognize that 
% machine conference does not frequently publish position papers: 
% most submissions consists of brief exposition, 
% a mathematical description of methodology, 
% and experimental findings. 
% Few papers are submitted without these components.
% However, we believe that critical writing offers 
% the best mechanism for addressing address problem formulation.
We argue that the paucity of critical writing in the machine learning community is problematic. 
When we have solid problem formulations,
flaws in methodology can be addressed 
by articulating new methods.
But when the problem formulation itself is flawed, 
neither algorithms nor experiments 
are sufficient to address the underlying problem.

Moreover, as machine learning continues 
to exert influence upon society, 
we must be sure that we are solving the right problems. 
While lawmakers and policymakers must increasingly 
consider the impact of machine learning,
the responsibility to account for the impact of machine learning 
and to ensure its alignment with societal desiderata 
must ultimately be shared by practitioners and researchers in the field. 
Thus, we believe that such critical writing ought to have a voice at machine learning conferences.

% \section{Acknowledgments}
% \label{sec:acknowledgment}
% \input{sections/acknowledgments.tex}

\bibliography{mythos}
\bibliographystyle{icml2017}

\end{document}